\pdfoutput=1
\documentclass[11pt]{article}

\usepackage[]{acl}

\usepackage{times}
\usepackage{latexsym}

\usepackage[T1]{fontenc}
\usepackage[utf8]{inputenc}

\usepackage{microtype}
\usepackage[normalem]{ulem}

\usepackage{inconsolata}

\usepackage{amsmath}
\usepackage{booktabs}
\usepackage{multirow}
\usepackage{array,multirow,graphicx}
\usepackage{float}
\usepackage{graphicx}
\graphicspath{ {/home/mlap/Research/google/mu-Plan/img/} }

\newcommand{\authorspace}[0]{\quad}

\usepackage{rotating}
\usepackage{accents}

\newcommand{\muplan}{\textsc{µplan}}
\title{\muplan: Summarizing using a Content Plan as Cross-Lingual Bridge}

\author{Fantine Huot \authorspace 
Joshua Maynez \authorspace
Chris Alberti \authorspace \\
\textbf{Reinald Kim Amplayo \authorspace Priyanka Agrawal \authorspace Constanza Fierro \authorspace} \\
\textbf{Shashi Narayan \authorspace Mirella Lapata} \\\\
Google DeepMind \\
\texttt{\small \{fantinehuot,joshuahm,chrisalberti,reinald,priyankagr,constanzam,shashinarayan,lapata\}@google.com}
}

\begin{document}
\maketitle
\begin{abstract}
Cross-lingual summarization aims to generate a summary in one language
given input in a different language, allowing for the dissemination of
relevant content among different language speaking populations.  The
task is challenging mainly due to the paucity of cross-lingual
datasets and the compounded difficulty of summarizing \emph{and}
translating.
This work presents \muplan, an approach to cross-lingual summarization that uses an intermediate planning step as a cross-lingual bridge. 
We formulate the plan as a sequence of entities capturing the
summary's content and the order in which it should be
communicated. Importantly, our plans abstract from surface form: using
a multilingual knowledge base, we align entities to their canonical
designation across languages and generate the summary conditioned on
this cross-lingual bridge and the input.\footnote{Source code and plan-annotated data are available at \url{https://github.com/google-deepmind/muplan}.}  Automatic and human evaluation on the  XWikis dataset (across four language pairs) demonstrates 
that our planning objective achieves state-of-the-art performance in
terms of informativeness and faithfulness. Moreover, \muplan\ models
improve the \emph{zero-shot} transfer to new cross-lingual language pairs
compared to baselines without a planning component.
\end{abstract}

\section{Introduction}

Given a document or multiple documents in a source language (e.g.,
English), cross-lingual summarization \cite{wang2022survey} aims to generate a summary in a
different target language (e.g., Czech or German). It enables the rapid dissemination of relevant
content across speakers of other languages. For instance, providing
summaries of English news articles to Czech or German speakers; or
making available to English speakers the content of product and
service descriptions in foreign languages.

\begin{figure*}[t]
    \centering
    \includegraphics[width=\textwidth,trim={50 140 27 125},clip]{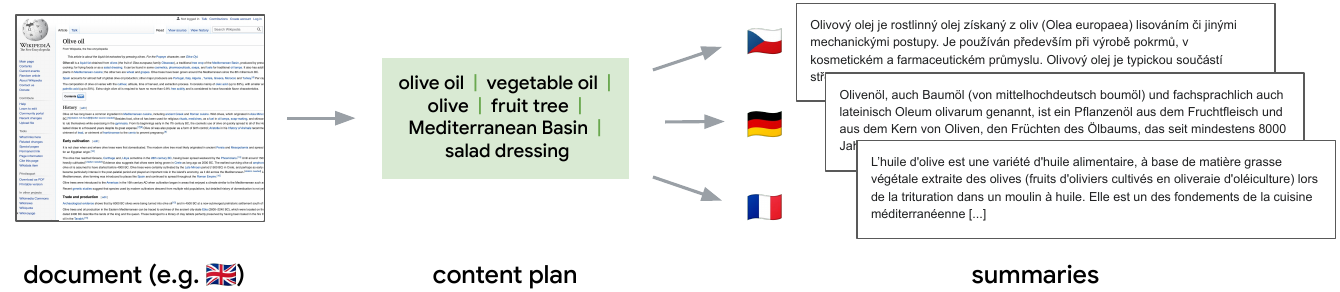}
    \caption{Source document and  content plan in English;
      target summaries in Czech, German, and French.}
    \label{fig:muplan}
\end{figure*} 

Recent years have seen tremendous progress in abstractive
summarization \cite{rush-etal-2015-neural,zhang2020pegasus} thanks to
advances in neural network models and the availability of large-scale
datasets
\cite{nytcorpus,hermann-nips15,grusky-etal-2018-newsroom}. While
initial efforts have focused on English, more recently, with the
advent of cross-lingual representations
\cite{crossling-ruder.et.al2019} and large pre-trained models
\cite{devlin-etal-2019-bert,liu-etal-2020-multilingual-denoising},
research on multilingual summarization (i.e.,~building monolingual
summarization systems for different languages) has also gained
momentum
\cite{chi2019crosslingual,scialom-etal-2020-mlsum,aharoni2022mface}.

Cross-lingual summarization faces the compounded challenge of having
to tackle difficulties relating to both monolingual summarization
(e.g.,~long inputs and outputs, hallucinations;
\citealt{maynez-etal-2020-faithfulness}) \emph{and} machine
translation (e.g.,~data imbalance, alignment across languages;
\citealt{koehn-knowles-2017-six}). Recent work has shown that
introducing an intermediate content planning step is helpful for
summarization in English, resulting in higher quality summaries,
especially in terms of faithfulness
\cite{narayan-etal-2021-planning,narayan2022conditional,huot-etal-2023-text}. In
this work, we argue that content planning also has the potential for
producing higher quality outputs for cross-lingual summarization.  In
particular, it provides a way of sharing task-specific knowledge
across languages, while formalizing important aspects of the
summarization task: identifying salient content in the source
documents, organizing this information in a meaningful order, and
standardizing it across different source and target language pairs.

We present \muplan, a cross-lingual summarization method that uses
content planning as a cross-lingual bridge
(Figure~\ref{fig:muplan}). Building upon previous work
\cite{narayan-etal-2021-planning}, we express our content plans as
entity chains, i.e.,~ordered sequences of salient entities. Although
more elaborate plan representations have been proposed in the
literature
\cite{wang-etal-2022-guiding,puduppully-etal-2022-data,narayan2022conditional},
entities are a natural choice for our task for two reasons.  They can
mitigate hallucinations in generated summaries which are commonly
related to entities
\cite{cao-etal-2022-hallucinated,zhao-etal-2020-reducing,maynez-etal-2020-faithfulness}
and are well-suited as a bridge across languages, thanks to the
availability of multilingual knowledge bases (e.g.,~DBpedia) which
represent entities in different languages. An interesting question for
our summarization task is which language to use for the content plan,
given that the source document and target summary are in different
languages. We employ a multilingual knowledge base to align the
entities across languages, which allows us to canonically transpose
the plan to different languages without the use of machine
translation.

\begin{figure*}[t]
    \centering
        \includegraphics[width=\textwidth,trim={5 250 80 0},clip]{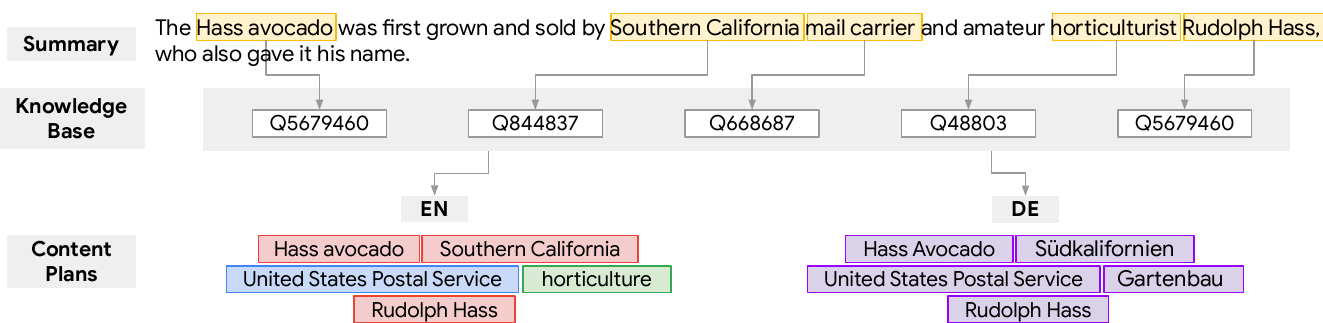}
    \caption{Plan annotation on an example summary (salient
      entities  highlighted in yellow). After pivoting on the
      knowledge base, corresponding canonical entities in English are
      shown in the bottom left. Most times they match the surface form
      in the summary (in red), other times they have the same root (in
      green) but they could differ greatly when entities need
      disambiguation (in blue). The aligned German content plan is
      shown in the bottom right.}
    \label{fig:plans_with_pivots}
\end{figure*}

We use a Transformer-based encoder-decoder model \cite{vaswani2017attention}
that first encodes the  document in the source language and then
decodes to generate  an intermediate plan representation and 
 the summary in the target language conditioned on the plan and the
input. We evaluate our method on the XWikis dataset
\cite{perez-beltrachini-lapata-2021-models}, a cross-lingual
abstractive summarization dataset derived from
Wikipedia\footnote{\url{https://www.wikipedia.org/}} articles aligned
across four different languages (English, Czech, French, and
German). We augment the training data for fine-tuning by annotating
each target summary with its corresponding content plan.

We investigate two distinct cross-lingual tasks, namely from English
to other languages ($\textsc{en}~\rightarrow ~\textsc{all}$) and from
other languages to English
($\textsc{all}~\rightarrow~\textsc{en}$). We demonstrate that models
fine-tuned with our planning objective outperform regular generated
summaries both in terms of ROUGE and faithfulness on the XWikis
dataset across all language pairs, in both settings.  Given the
scarcity of cross-lingual datasets, we also investigate zero-shot
cross-lingual transfer to new language pairs and demonstrate that
\muplan\ models outperform comparison systems without planning
components. 

Our contributions can be summarized as follows: (a)~we introduce
a training objective for cross-lingual abstractive summarization that
uses \textbf{entity planning as a bridge between languages}. Using automatic and human evaluation, we
 show that it yields better quality summaries and more
effective zero-shot transfer to new language pairs than non-planning
baselines; and (b)~we leverage a multilingual knowledge base to
annotate the training data with plans, thus \textbf{transposing entity
  names to their canonical designation} in all languages, avoiding
errors induced by mistranslation altogether. This  strategy
enables the mapping of entities that do not have an equivalent name in
the target language to fully-localized paraphrases.

\section{Related Work}

\paragraph{Cross-lingual Summarization} A key challenge in
cross-lingual summarization is the scarcity of training data.  Indeed,
while creating large-scale multilingual summarization datasets has
proven feasible
\cite{straka-etal-2018-sumeczech,scialom-etal-2020-mlsum}, naturally
occurring documents in a source language paired with summaries in
different target languages are rare.  For this reason, existing
cross-lingual approaches create large-scale synthetic data using
machine translation
\cite{zhu-etal-2019-ncls,cao-etal-2020-jointly,ouyang-etal-2019-robust}.

Cross-lingual benchmarks include WikiLingua
\cite{ladhak-etal-2020-wikilingua}, a dataset derived from
multilingual how-to guides, which  are relatively short and
their summaries  limited to brief instructional sentences. CrossSum
\cite{bhattacharjee2021crosssum} contains over a million article and
summary samples, aligned from the multilingual XL-Sum
\cite{hasan2021xl} dataset, but the summaries are limited to one or
two sentences.  \citet{fatima-strube-2021-novel} propose a Wikipedia-based cross-lingual dataset, but it only includes the English to German language direction. We work with XWikis
\cite{perez-beltrachini-lapata-2021-models}, a cross-lingual dataset
derived from Wikipedia with long input documents and long target
summaries across four languages: English, Czech, French, and German. We compare these datasets in 
Appendix~\ref{sec:crosslingual-datasets}.

\paragraph{Content Plans for Summarization} The idea of breaking down
the generation task into smaller steps through a separate planning
stage has proven helpful for data-to-text generation
\citep{puduppully2019data,moryossef-etal-2019-step,puduppully2021data,liu-chen-2021-controllable}
and lately for summarization and long-form question
answering~\citep{narayan-etal-2021-planning,narayan2022conditional}.
Our work is closest to~\citet{narayan-etal-2021-planning} who show
that an intermediate planning step conceptualized as a sequence of
salient entities could yield more faithful and entity-specific
summaries.  Herein, we explore whether  content plans can serve
as a cross-lingual bridge and enable \emph{task transfer} between languages.

\paragraph{Zero-shot Cross-lingual Transfer}
A substantial portion of the work on zero-shot cross-lingual transfer
has focused on classification tasks~\citep{extreme-pmlr-2020}, such as
XNLI~\citep{10.1162/tacl_a_00288}, part-of-speech tagging, dependency
parsing, named entity recognition~\citep{ansell-etal-2021-mad-g}, and
question answering~\citep{conneau-etal-2020-unsupervised}. Some recent
work has also investigated generative tasks in the zero-shot
setting. \citet{johnson-etal-2017-googles} show that by prepending a
special token to the input text to indicate the target language of the
translation, models learn to perform implicit bridging between
language pairs unseen during training.  \citet{chen-etal-2021-zero}
perform zero-shot cross-lingual machine translation, by using parallel
data in only one language pair and leveraging a multilingual encoder
to support inference in other languages.
\citet{vu-etal-2022-overcoming} study how to fine-tune language models
on only one language to perform zero-shot cross-lingual summarization
in other languages, by adding unlabeled multilingual data.
\citet{whitehouse-etal-2022-entitycs} use Wikidata to improve
zero-shot cross-lingual transfer for code-switching in a number of
entity-centric downstream tasks. We also resort to Wikidata to obtain
a canonical designation of entities across languages, however, the use
of plans as a cross-lingual bridge for summarization is new to our knowledge.

\section{Plans as a Cross-Lingual Bridge}

\subsection{Problem Formulation}

We formalize the cross-lingual abstractive summarization task as
follows: Given an input document~$d$ in a source
language~\textsc{src}, generate a summary~$s$ in target
language~\textsc{tgt}. We model this as $p(s|d)$.

For the content planning objective, our goal is to teach the model to
first generate a content plan~$c$ for the summary as~$p(c|d)$, before
generating the summary itself as~$p(s|c,d)$. Following
\citet{narayan-etal-2021-planning}, instead of modeling $p(c|d)$ and
$p(s|c,d)$ separately, we train the model to generate the concatenated
plan and summary sequence $c;s$. As a result, the model first
generates the content plan $c$ and then continues to generate the
summary~$s$ conditioned on both~$c$ and~$d$. In the following section,
we describe how we annotate the data with content plans for this
planning objective.

\begin{table*}[t]
    \centering
    \begin{tabular}{lp{0.44\textwidth}p{0.37\textwidth}}
        \toprule
         & \multicolumn{1}{c}{Summary}  & \multicolumn{1}{c}{Plan}  \\
        \midrule
        \small $\textsc{en}\rightarrow\textsc{cs}$ & \small Richard Dagobert Brauer byl \textcolor{blue}{německý} \textcolor{red}{matematik} žijící v \textcolor{magenta}{USA}. Pracoval zejména v oblastech abstraktní \textcolor{orange}{algebry} a \textcolor{olive}{teorie čísel}. Je také zakladatelem modulární teorie reprezentací. & \small \textcolor{blue}{German Empire \& Německé císařství} | \textcolor{red}{mathematician \& matematik} | \textcolor{magenta}{United States of America \& Spojené státy americké} | \textcolor{orange}{algebra \& algebra} | \textcolor{olive}{number theory \& teorie čísel} \\
        \small $\textsc{en}\rightarrow\textsc{fr}$ &  \small CALET est un \textcolor{blue}{observatoire spatial} développé par le \textcolor{red}{Japon} et installé en 2015 à bord de la \textcolor{magenta}{Station spatiale internationale}. Cet instrument analyse les \textcolor{orange}{rayons cosmiques} et le \textcolor{olive}{rayonnement gamma} à haute énergie avec comme objectif principal l'identification des éventuelles signatures de la \textcolor{cyan}{matière noire}. & \small \textcolor{blue}{space observatory \& télescope spatial} | \textcolor{red}{Japan \& Japon} | \textcolor{magenta}{International Space Station \& station spatiale internationale} | \textcolor{orange}{cosmic radiation \& rayonnement cosmique} | \textcolor{olive}{gamma ray \& rayon gamma} | \textcolor{cyan}{dark matter \& matière noire} \\
        \small $\textsc{de}\rightarrow\textsc{en}$ & \small The TKS spacecraft ("Transport Supply Spacecraft", \textcolor{blue}{GRAU} index 11F72) was a \textcolor{red}{Soviet} \textcolor{magenta}{spacecraft} conceived in the late 1960s for resupply flights to the military \textcolor{orange}{Almaz} space station. 
 & \small{\textcolor{blue}{Hauptverwaltung für Raketen und Artillerie \& GRAU} | \textcolor{red}{Sowjetunion \& Soviet Union} | \textcolor{magenta}{Raumschiff \& spacecraft} | \textcolor{orange}{Almas \& Almaz}} \\
        \bottomrule
    \end{tabular}
    \caption{Summaries with annotated 
      plans. Same color denotes alignment between entities in the plan and 
       summary. Plans are entities in the language of the source document \emph{and} (diacritic \&) the language of the target summary.}
    \label{tab:annotation_examples}
\end{table*}

\subsection{Content Plans}

Similarly to \citet{narayan-etal-2021-planning}, we formulate the content plan as an ordered sequence of entities. Figure~\ref{fig:plans_with_pivots} illustrates our  annotation process.  We annotate each example with its corresponding content plan by extracting  salient entities, i.e.,~entities that are important to mention when summarizing. 

We extend this paradigm by linking each entity to its entry in a
multilingual knowledge base. This way we obtain a canonical
designation of each entity, removing morphology and selecting the most
common designation out of multiple aliases. The knowledge base also
provides disambiguation when it is needed.  We use entity names in
the content plans, instead of knowledge base indices, in order to
leverage the natural language capabilities of pretrained language
models.

We then use the inter-language information from the knowledge base to
pivot content plans across languages. For each entity, we obtain
its canonical designation in both the language of the source document
and the language of the target summary.  We provide an example of the multilingual
mappings in our annotated content plans in
Figure~\ref{fig:plans_with_pivots}. 

This strategy enables the mapping of entities that do not have an equivalent name in the target language to fully-localized names. And the model learns to generate a content plan of localized entities, avoiding errors induced by translation. 

Finally, we compose the content plan as a sequence of canonical entity
names, each expressed in pairs in both the source and target language
(Table \ref{tab:annotation_examples}).  We designate the planning
objective using these cross-lingual content plans as \muplan.

\subsection{Summarization Tasks}
We next define the summarization tasks considered in this
work, and our assumptions about the   cross-lingual
training data being available. 
\paragraph{Cross-Lingual Tasks}
In what follows, let $\mathcal{L}$ be the set of all languages, \textsc{src} the language of the source document, and \textsc{tgt} the language of the target summary. We denote the cross-lingual data as
$\mathcal{D}_{\textsc{src}\rightarrow \textsc{tgt}}$, e.g.,
% $\mathcal{D}_{\textsc{src\_tgt}$, e.g.,
$\mathcal{D}_{\textsc{en}\rightarrow \textsc{cs}}$ for Czech summaries aligned with English inputs. Analogously, we denote the monolingual data as $\mathcal{D}_{\textsc{lang}}$, e.g., $\mathcal{D}_{\textsc{cs}}$ for Czech summaries with Czech inputs.

Herein, we investigate two specific cross-lingual tasks: (a)~from
English to other languages and (b)~from other languages to English,
which we denote as $\textsc{en}\rightarrow \textsc{all}$ and
$\textsc{all}\rightarrow \textsc{en}$, respectively. The
$\textsc{en}\rightarrow \textsc{all}$ task is the main focus of our
work. The task is particularly interesting because it would make a
large amount of English information available to speakers of other
languages but also challenging since it involves a cross-lingual
summarization model that can generate fluent text in many languages.
We define the data for the  $\textsc{en}\rightarrow \textsc{all}$ task as: 
\[
\mathcal{D}_{\textsc{en}\rightarrow \textsc{all}} = \mathcal{D}_\textsc{en} \cup \bigcup_{\textsc{tgt} \in
  \mathcal{L} - \{\textsc{en}\}} \mathcal{D}_{\textsc{en}\rightarrow \textsc{tgt}},
\]
and for the $\textsc{all}\rightarrow \textsc{en}$, task as:
\[
\mathcal{D}_{\textsc{all}\rightarrow \textsc{en}} = \mathcal{D}_\textsc{en} \cup \bigcup_{\textsc{src} \in
  \mathcal{L} - \{\textsc{en}\}} \mathcal{D}_{\textsc{src}\rightarrow \textsc{en}}.
\]
Note that both tasks have access to monolingual
\textsc{en} data.  For models that do not use an intermediate planning
step, each data example is a document and summary pair $(d,s)$. For
\muplan\ models, each data example also includes a content plan, $(d,c;s)$.

\paragraph{Zero-Shot Cross-Lingual Tasks} 
Given the scarcity of cross-lingual datasets, we investigate whether
\muplan~ can help with zero-shot cross-lingual transfer to new
language pairs. For each target language \textsc{tgt}, we perform
zero-shot transfer experiments on the $\textsc{en}\rightarrow
\textsc{all}$ task by holding out the $\textsc{en}\rightarrow
\textsc{tgt}$ cross-lingual data during fine-tuning. We then evaluate
performance on the $\textsc{en}\rightarrow \textsc{tgt}$ test data. To
ensure that the model maps the language token to the correct language
and to prevent catastrophic forgetting of the \textsc{tgt} language
during fine-tuning \cite{vu-etal-2022-overcoming}, we include 
\textsc{tgt} monolingual summarization data in the fine-tuning data
mixture, under the assumption that  monolingual data is easier to
come by than cross-lingual data. We denote this zero-shot
cross-lingual transfer task as $\textsc{en}\rightarrow
\textsc{tgt}_\textsc{zs}$ and define~as:
\[
\mathcal{D}_{\textsc{en}\rightarrow \textsc{tgt}_\textsc{zs}} = \mathcal{D}_\textsc{en} \cup \mathcal{D}_\textsc{tgt} \cup\bigcup_{\textsc{l} \in
  \mathcal{L} - \{\textsc{en,tgt}\}} \mathcal{D}_{\textsc{en}\rightarrow \textsc{l}}.
\]
For greater generalization, we could use unlabeled monolingual data (without summaries), however, we leave this to future work. 

\section{Experimental Setup}

\subsection{Dataset} The XWikis dataset
\cite{perez-beltrachini-lapata-2021-models} was created from Wikipedia
articles under the assumption that the body and lead paragraph
constitute a document-summary  pair.  Cross-lingual  document-summary
instances  were  derived by combining lead paragraphs and articles’
bodies from language-aligned Wikipedia titles. Although XWikis covers
only four  languages, English (\textsc{en}), Czech (\textsc{cs}),
German (\textsc{de}), and French (\textsc{fr}),  the  dataset creation
procedure  is  general  and  applicable  to any  languages
represented in  Wikipedia. 

\setlength{\tabcolsep}{8pt}
\begin{table}[t]
    \centering
\small
    \begin{tabular}{lrrl}
        \toprule
        & \multicolumn{1}{c}{Train}  & \multicolumn{1}{c}{Validation}    & \multicolumn{1}{c}{Test} \\
        \midrule
        \textsc{en}     & 624,178   & \hspace*{-1cm}8,194     & 7,000 \\
        $\textsc{en}\rightarrow \textsc{cs}$ & 134,996   & 250$^{\dagger}$     & 6,855$^{\dagger}$ \\
        $\textsc{en}\rightarrow \textsc{de}$ & 409,012   & 250$^{\dagger}$    & 9,750$^{\dagger}$ \\
        $\textsc{en}\rightarrow \textsc{fr}$ & 451,964   & 250$^{\dagger}$     & 9,727$^{\dagger}$ \\
        $\textsc{cs}\rightarrow \textsc{en}$ & 48,519    & 2,549 & 6,999 \\
        $\textsc{de}\rightarrow \textsc{en}$ & 344,438   & 18,160 & 6,999 \\
        $\textsc{fr}\rightarrow \textsc{en}$ & 283,182   & 14,899  &  6,992 \\
        \bottomrule
    \end{tabular}
    \caption{Number of data samples in the XWikis dataset and  splits considered in this work. New splits for  the  $\textsc{en}\rightarrow \textsc{all}$ 
      language pairs are
      marked by $^{\dagger}$.} 
    \label{tab:xwikis}
\end{table} 

Table~\ref{tab:xwikis} shows the number of data samples for each
language pair. Note that the $\textsc{en}\rightarrow \textsc{tgt}$
language pairs are not parallel between all languages. Cross-lingual
language pairs in the $\textsc{all}\rightarrow \textsc{en}$ setting
have separate training, validation and test splits, but in the
$\textsc{en}\rightarrow \textsc{all}$ setting there are only training
and validation splits. Therefore, for all the $\textsc{en}\rightarrow
\textsc{all}$ cross-lingual language pairs, we separate the validation
split into two, taking the first 250 examples for validation and the
rest for testing.

The XWikis dataset provides the input documents as a list of section titles and paragraphs that constitute the body of the Wikipedia article to summarize. We format the input documents by concatenating the titles and paragraphs, marking each title with an end-of-title token \textsc{eot} and each paragraph with an end-of-paragraph token \textsc{eop}. We prepend the source language code and target language code to the input document for each cross-lingual document and summary pair.

Since the XWikis dataset is derived from Wikipedia, we annotate the plans by extracting all the entities from the reference summaries that have embedded hyperlinks. We then exclude the ones that correspond to phonetic pronunciations. For each of the remaining hyperlinks, we query the Wikidata knowledge base\footnote{\url{https://www.wikidata.org/}} to extract the ID of the entity (e.g., ‘Q844837’) corresponding to the hyperlink URL (e.g., \url{https://en.wikipedia.org/wiki/Southern_California}). Querying  Wikidata again for this entity ID allows us to retrieve its  canonical name  in different languages (e.g., ‘Southern California’ in English, or ‘Südkalifornien’ in German; see Figure~\ref{fig:plans_with_pivots}).
The XWikis dataset was generated from a 2016 Wikipedia data dump and we used one from 2023 for extracting the hyperlinks from the summaries. Therefore, for articles that went through significant changes between 2016 and 2023, the pages were not aligned and we did not annotate these examples with content plans. 
This problem affects about 4.5\% of the training data.
We create a \emph{filtered} version of the training data that excludes these examples with missing content plans.

\begin{table*}[t]
\small
  \centering
    \begin{tabular}{@{}lp{6cm}p{6cm}@{}}
        \toprule
        \multicolumn{1}{c}{Plan Type} & \multicolumn{1}{c}{Predicted Plan}  & \multicolumn{1}{c}{Gold Plan}  \\
        \midrule
        \textsc{src[en]}     & Dutch | fortification | Banda Neira | Maluku Islands | Netherlands | Dutch East Indies   &  Banda Neira | Banda Islands | Maluku Islands | Indonesia | Maluku | nutmeg  \\
          \textsc{tgt[de]} & Estland | Folk Metal | Band | Tallinn | Markus Lõhmus & Estland | Folk Metal | Euphemismus | Wolf \\
        \textsc{src[en]\_tgt[fr]} & county seat \& siège de comté | Crawford County \& comté de Crawford | Arkansas \& Arkansas | United States of America \& États-Unis  
 & Arkansas \& Arkansas | United States of America \& États-Unis \\

        \bottomrule
    \end{tabular}
    \caption{Examples of generated and gold content plans for different source and target languages.}
    \label{tab:plan_examples}
\end{table*} 

\subsection{Comparison Models}

We demonstrate \muplan~on both the $\textsc{en}\rightarrow \textsc{all}$ and $\textsc{all}\rightarrow \textsc{en}$ tasks and  compare it with a number of different modeling approaches.

\paragraph{Machine Translation} A common approach is to adopt a machine translation-based pipeline which can be used in two ways:
(a) first translate the original document into the target language and
then summarize the translated document or (b) first summarize the
original document and then translate the summary
\cite{ouyang-etal-2019-robust,wan-etal-2010-cross,ladhak-etal-2020-wikilingua}. We
denote the former approach as Translate-train (TR$_{train}$) and the
latter as Translate-test (TR$_{test}$). We perform  machine
translation with Google Translate.

Previous work
\cite{kramchaninova-defauw-2022-synthetic,vu-etal-2022-overcoming} has
highlighted various limitations with these approaches such as  dependence on the quality of available machine translation
systems in a given language and in turn the availability of
high-quality parallel data, a potential misalignment
of the data after translation, and translationese artifacts
\cite{clark-etal-2020-tydi}. 

\paragraph{End-to-end Summarization} 
This approach, which we denote as \textsc{e2e}, directly
fine-tunes a multilingual pretrained model on the cross-lingual data
\cite{perez-beltrachini-lapata-2021-models}. It does not incorporate a
planning component, but avoids the potential error propagation problem
of machine translation pipeline systems.
 
\paragraph{\muplan\ Variants} 
% Explain different plans

We experiment with different plan formulations to establish which type
of plan performs well as a cross-lingual bridge. The language of the
source document being different from the language of the target
summary raises the question of which language to use for the content
plans. In the default \muplan\ setup, entities in the plan are
expressed in pairs, with their canonical name in both the language of
the source document and the language of the target summary. In
addition, we explore two alternatives: (a)~entity names only in the
source language and (b)~entity names only in the target language.
Table~\ref{tab:plan_examples} presents examples of different language plans. 
Moreover, we experiment with the internal constitution of the plans: we provide the length of the gold plan during training [$\textsc{length}$], and shuffle entities to investigate the importance of the sequence order [$\textsc{shuffle}$]. Since the quality of the plan annotations is dependent on the quality of the entity linking, we also investigate the impact of partially corrupted gold plans, by dropping a portion of the plan entities at random during training. We denote these experiments as [\textsc{corrupt20}] and [\textsc{corrupt30}], in which we drop 20\% and 30\% of the entities, respectively.

\paragraph{Model Training} 

All baselines and \muplan\ variants are based on the mT5 model
(\citealt{xue-etal-2021-mt5}; XL 3.7B parameters) which we finetune
with maximum input and output sequence lengths of~2,048 and
256~tokens, respectively. Our models are finetuned on Cloud TPU v3 with a learning
rate of~0.002, a batch size of~128, up to 80,000~steps, evaluating
every~1,000 steps. We select the best checkpoints by measuring ROUGE-L (see Section~\ref{sec:automatic-eval} for details)
on 250~examples of the validation split for each language pair and
take the best unweighted average across all language pairs.

\paragraph{Note on LLMs} We performed few-shot experiments with LLMs, however, these were consistently  inferior to our fine-tuned systems  confirming the observations of \citet{maynez2023benchmarking}. It is particularly challenging to learn to plan and summarize simply from a few examples. We report LLM experiments (1-shot, no planning) in Appendix~\ref{sec:few-shot-LLM}. 

\section{Results}
\label{sec:results}

\subsection{Automatic Evaluation}
\label{sec:automatic-eval}

\begin{table*}[t]
  \centering
%   \small
  \begin{tabular}{l|cccc|cccc}\toprule
 \multicolumn{1}{c}{}   &   \multicolumn{4}{c}{ROUGE-L} & \multicolumn{4}{c}{XNLI} \\
  & \multicolumn{1}{c}{\textsc{TR}$_{train}$} &
  \multicolumn{1}{c}{\textsc{TR}$_{test}$} &
  \multicolumn{1}{c}{\textsc{e2e}} & \multicolumn{1}{c}{\muplan}
&  \multicolumn{1}{|c}{\textsc{TR}$_{train}$} &
  \multicolumn{1}{c}{\textsc{TR}$_{test}$} &
  \multicolumn{1}{c}{\textsc{e2e}} & \multicolumn{1}{c}{\muplan}  \\ \midrule
 \small{$\textsc{en}\rightarrow \textsc{en}$}    & 37.42 & 37.38 & 37.57  & \textbf{39.53}   & 53.99  & 47.50 & 53.54 & \textbf{56.16}\\
  \small{$\textsc{en}\rightarrow \textsc{cs}$}    & 32.81 & 26.26 & 32.74  & \textbf{33.18}   & 34.32  & 36.90 & 33.79 & \textbf{37.70}\\
  \small{$\textsc{en}\rightarrow \textsc{de}$}    & 38.28 & 28.47 & 38.58  & \textbf{38.94}   & 39.52  & 38.19 & 38.92 & \textbf{42.98}\\
  \small{$\textsc{en}\rightarrow \textsc{fr}$}    & 41.19 & 31.59 & 41.36  & \textbf{41.57}   & 41.45  & 40.75 & 40.83 & \textbf{52.72}\\ \midrule
  \textbf{$\textsc{en}\rightarrow \textsc{all}$}  & \uline{37.42} & \uline{30.93} & \uline{37.56}  & \textbf{38.30}   & \uline{42.32}  & \uline{40.84} & \uline{41.77}      & \textbf{47.39} \\ \bottomrule
\multicolumn{1}{c}{}\\ \toprule
 \multicolumn{1}{c}{}   &   \multicolumn{4}{c}{ROUGE-L} & \multicolumn{4}{c}{XNLI} \\
& \textsc{TR}$_{train}$ & \textsc{TR}$_{test}$ & \textsc{e2e} & \muplan
& \textsc{TR}$_{train}$ &\textsc{TR}$_{test}$ & \textsc{e2e} & \muplan
  \\ \midrule
\small{$\textsc{en}\rightarrow \textsc{en}$} & 33.15 & 34.43 & 35.47 
& \textbf{36.09} & 63.29 & \textbf{66.46} & 51.79 & 60.71\\
\small{$\textsc{cs}\rightarrow \textsc{en}$} & 29.47 & 31.93 & \textbf{33.30} 
& 32.82 & \textbf{45.39} & 30.39 & 30.14 &  30.81\\
\small{$\textsc{de}\rightarrow \textsc{en}$} & 29.89 & 32.48 & 33.70
& \textbf{34.32} & \textbf{45.20} & 42.17 & 35.22 & 41.16\\
\small{$\textsc{fr}\rightarrow \textsc{en}$} & 29.60 & 32.35 & 33.22
& \textbf{34.20} & \textbf{41.63} & 39.81 & 32.58 & 39.34\\\midrule
\textbf{$\textsc{all}\rightarrow \textsc{en}$} & \uline{30.53} & \uline{32.80} & \uline{33.92}
& \textbf{34.36} & \uline{\textbf{48.88}} & \uline{44.71} & \uline{37.43} & 43.00                       \\ \bottomrule
\end{tabular}
    \caption{ROUGE-L and XNLI  results per  language pair and overall
      for  the $\textsc{en}\rightarrow \textsc{all}$ and
      $\textsc{all}\rightarrow \textsc{en}$ tasks. Systems significantly different from \muplan\ are \uline{underlined} (using paired bootstrap resampling; \mbox{$p < 0.05$}).}
    \label{tab:results_per_language}
\end{table*}

We automatically evaluate system output along the 
dimensions of  summary relevance, summary faithfulness, and content plan
relevance.  For \emph{summary relevance}, we use ROUGE \cite{lin-2004-rouge}
to compare system-generated summaries with gold-standard ones. Since
the availability of word tokenizers differs for non-English languages,
we follow  \citet{aharoni2022mface} and
compute ROUGE with 
a SentencePiece tokenizer \cite{kudo-richardson-2018-sentencepiece}
trained on mC4 \cite{xue-etal-2021-mt5}.

In terms of \emph{summary faithfulness}, following
\citet{honovich-etal-2022-true-evaluating}, we employ an entailment
classifier that predicts whether the input document supports the
output summary. In line with previous work
\cite{narayan2022conditional,schuster-etal-2022-stretching}, we split
the summary into sentences for a more fine-grained evaluation. We
predict the entailment of each sentence and average the entailment
scores. We use an mT5-XXL model \cite{xue-etal-2021-mt5}
trained on XNLI \cite{conneau-etal-2018-xnli}, a multilingual NLI
dataset. There are currently no cross-lingual datasets for NLI,
however our preliminary analysis reported in
Appendix~\ref{sec:nli-analysis} shows that an XNLI-trained mT5 model
works well in predicting cross-lingual entailment. It has the added
benefit of avoiding potential error propagation from introducing a
machine translation step in the evaluation process (e.g., translating
the document or the summary in English).  Finally, we evaluate \emph{plan relevance}, by comparing 
generated content plans against  gold-standard ones. Specifically,
we compute F1 scores on the entities in the predicted summaries
against the corresponding reference entities. 

\paragraph{Planning outperforms translation-based approaches}
Table \ref{tab:results_per_language} presents an overview of our
results for the $\textsc{en}\rightarrow \textsc{all}$ and
$\textsc{all}\rightarrow \textsc{en}$ tasks. We report results on the
filtered data, as we observed little difference overall between
filtered and non-filtered training samples (results with non-filtered
training data are provided in Appendix \ref{sec:filtered}). Moreover,
for the sake of brevity, we only present ROUGE-L results, however see
Appendix~\ref{sec:more-results} for additional metrics.
We see that \muplan\ consistently outperforms both the
translation-based approaches and the non-planning baseline
(\textsc{e2e}) in terms of ROUGE-L and XNLI scores on both
$\textsc{en}\rightarrow \textsc{all}$ and $\textsc{all}\rightarrow
\textsc{en}$ tasks.  Note that TR$_{train}$ is the overall winner according 
to XNLI in the $\textsc{all}\rightarrow
\textsc{en}$ task. We hypothesize that the higher XLNI scores for TR$_{train}$ are to some extent an artifact of translation and the XNLI model. Indeed, machine translation tends to drop information during the translation process, which biases TR$_{train}$ towards higher XNLI scores. The other reason is that the XNLI model itself has been trained on more English data and just works better in this setting as it is faced with a simpler monolingual task (both the input document and summary are in English). Previous work \cite{perez-beltrachini-lapata-2021-models} has focused on 
$\textsc{all}\rightarrow
\textsc{en}$  tasks using   m\textsc{Bart}50 
\cite{tang2020multilingual} and \textsc{e2e} models; they report an average ROUGE-L of 32.76 for the same language pairs shown in Table~\ref{tab:results_per_language} (last row). 

\begin{table}[t]
\centering
  \resizebox{0.8\columnwidth}{!}{%
    \begin{tabular}{@{}l@{~}ccc@{}}
        \toprule
        & ROUGE-L  & XNLI & F1  \\
        \midrule
        \muplan &  38.30 &	47.39   &  0.40  \\
        \muplan$_{\textsc{src}}$    & 38.14 &	 47.72   &  0.41  \\
        \muplan$_{\textsc{tgt}}$ & 37.97 &	47.37   & 0.40  \\
        \midrule
        \muplan$_{\textsc{length}}$ &  37.09 &	45.71 & 0.37  \\
        \muplan$_{\textsc{shuffle}}$ &  38.01	& 46.25 & 0.40  \\
        % \muplan$_{\textsc{union}}$ &  38.00 &	45.82 &0.22  \\
        \muplan$_{\textsc{corrupt20}}$ &  38.34 & 47.46 & 0.33  \\
        \muplan$_{\textsc{corrupt30}}$ &  38.17	& 46.55 & 0.30  \\        
        \midrule
        \muplan$^{oracle}$ & 48.28 & 40.83	   & 1.00 \\
        \muplan$_{\textsc{src}}^{oracle}$    & 47.96 &	41.22   & 1.00    \\
        \muplan$_{\textsc{tgt}}^{oracle}$ & 48.13 & 40.84 & 1.00     \\
        \bottomrule
    \end{tabular}%
    }
    \caption{Comparison of different \muplan\ plan formulations (including oracles) on the $\textsc{en}\rightarrow \textsc{all}$ task.}
    \label{tab:plan_types}
\end{table} 

\paragraph{Best plans include entities in  source and target language} We compare different types of plan formulations on the $\textsc{en}\rightarrow~\textsc{all}$ task and report our results in Table~\ref{tab:plan_types}. Mixed language plans that contain entities in both the source and target language, which is the default \muplan\ setting, deliver better results than plans with entities in only one language (marked here as \textsc{src} and \textsc{tgt}). Table~\ref{tab:plan_examples} shows some plans generated by \muplan\ under these different settings and compares them to the gold ones. 

Predicted and gold plans have similar length, measured by the number of entities in the plan (6~on average). We also find  that gold and predicted plans have overlapping but not identical entities (the 
F1 score is around~0.4; see Tables~\ref{tab:plan_types} and~\ref{tab:plan_examples}). However, we do not expect perfect overlap;  gold summaries in
XWikis are derived from  lead paragraphs in Wikipedia articles, and as a result
some of the entities in the gold plans might not even appear in the source
document. This is corroborated by XNLI scores which are lower for oracle summaries compared to machine-generated ones.  Providing information about the length of the gold plan during
training, reported as \textsc{length}, does not affect the results
very much and actually yields slightly lower metrics than the default
\muplan\ setup. The \textsc{shuffle} metrics, for which the entity
order is shuffled, are similar to the default
setup. This result indicates that the order of the entities does not matter much
for planning the summary generation. 

The experiments with corrupted entity plans mimic the effects of an imperfect entity linking. At training time, we drop a percentage of the entities in the plan at random, denoted as \textsc{corrupt20} and \textsc{corrupt30}, for 20\% and 30\%, respectively. 
We observe that \muplan~is robust to some degree of noise in the plan annotation process, as there is only a slight decrease in  ROUGE-L and XNLI scores as the percentage of corruption increases.

\paragraph{Oracle plans show there is room for improvement}
For comparison, we report results when models have access to oracle
content plans, which we denote as \emph{oracle}. At inference time,
the encoder first encodes the source document, while the decoder gets
the gold plan as a forced prompt before generating the summary. These
oracle experiments provide an upper bound of how \muplan\ models would
perform in a best case scenario. In Table~\ref{tab:plan_types}, we see
that the oracle metrics are higher by a wide margin, of around 10
\mbox{ROUGE-L} points, from the best predicted results. This behavior
is expected and shows that models can correctly generate summaries  from plans in the target language but also from
aligned English plans. Moreover, these results  confirm that \muplan's mixed
language plans provide additional information that models can leverage
effectively. 

While  ROUGE-L scores are much better, we note that oracle plan experiments obtain lower XNLI scores overall. This behavior is somewhat expected since the XWikis dataset was created by associating the leading paragraph of a Wikipedia page with the body of the article. \citet{perez-beltrachini-lapata-2021-models}
verified whether  the lead paragraph constitutes a valid summary, by asking native speakers to ascertain for each sentence in the summary whether it was supported by the document. Overall, human judges viewed the summaries as an acceptable (but not perfect) overview of the Wikipedia document, with 60\%--78\% of the summary sentences being supported by the document, depending on language pairs.

\begin{table}[t]
  \centering
  \resizebox{0.8\columnwidth}{!}{%
  %\small
    \begin{tabular}{@{}l@{~}cc|cc@{}}
        \toprule
        &   \multicolumn{2}{c}{ROUGE-L}  & \multicolumn{2}{c}{XNLI} \\
& \textsc{e2e} & \muplan & \textsc{e2e} & \muplan\\ \midrule
    
        ${\textsc{en}\rightarrow~\textsc{cs}_{\textsc{zs}}}$ & 15.10 &  \bf 18.64 & 34.95 & \bf 39.04 \\
         ${\textsc{en}\rightarrow~\textsc{de}_{\textsc{zs}}}$ & 17.50 & \bf 19.18 & 45.51 & \bf 48.80 \\
         ${\textsc{en}\rightarrow~\textsc{fr}_{\textsc{zs}}}$ & 18.54 & \bf 23.61 & 45.51  & \bf 45.96 \\
     \bottomrule
    \end{tabular}%
    }
    \caption{Zero-shot cross-lingual transfer results.}
    \label{tab:zero_shot}
\end{table} 

\paragraph{Planning enables zero-shot transfer}
Table~\ref{tab:zero_shot} shows the results of our zero-shot
cross-lingual transfer experiments. We observe that \muplan\ delivers
higher ROUGE-L and XNLI scores when evaluated on an unseen language pair. This
indicates that an intermediate planning step helps transfer task
knowledge to new language pairs.

\begin{table}[t]
  \centering
      \resizebox{0.8\columnwidth}{!}{%
    \begin{tabular}{@{}l@{~~}r@{~~}r|r@{~~}c@{}}
        \toprule
        &   \multicolumn{2}{c}{ROUGE-L}  & \multicolumn{2}{c}{XNLI} \\
& \textsc{e2e} & \muplan & \textsc{e2e} & \muplan\\ \midrule
    
        $\textsc{en}\rightarrow~\textsc{all}$ & 9.15 &  \bf 9.33 & 31.38 & \bf 43.53 \\
         $\textsc{en}\rightarrow~\textsc{fr}$ & 22.03 & \bf 23.10 & 33.39 & \bf 47.63 \\
     \bottomrule
    \end{tabular}%
    }
    \caption{Zero-shot domain transfer results (CrossSum).}

    \label{tab:crosssum}
\end{table} 

\paragraph{Planning enables domain transfer}
In addition to these zero-shot cross-lingual transfer experiments, we extend our analysis to zero-shot domain transfer by applying the trained models on data from another domain. For this experiment, we select the CrossSum dataset \cite{bhattacharjee2021crosssum}, a cross-lingual dataset with article-summary pairs derived from news articles. While CrossSum summaries are much shorter than the XWikis ones and do not necessarily call for an intermediate planning step for content selection and organization, previous experiments show that \muplan\ brings improvements in faithfulness that might benefit CrossSum as well. We run inference on the test splits of CrossSum with the \textsc{e2e} and \muplan\ models trained on the XWikis corpus and report results in Table~\ref{tab:crosssum}. We observe that the \muplan\ model yields much better XNLI scores for comparable ROUGE-L scores, compared to the \textsc{e2e} model without planning. ROUGE-L scores are overall low for both models because for many language pairs, the models exhibit catastrophic forgetting due to the mismatch of languages between the CrossSum and the XWikis datasets. When inspecting the \textsc{en}$\rightarrow$ \textsc{fr} direction, which is present in both XWikis and CrossSum, we observe that \muplan\ brings improvements in both ROUGE-L and XNLI scores. 

\subsection{Human Evaluation}
\label{sec:human-evaluation}

In addition to automatic metrics, we also conducted a judgment
elicitation study.

Specifically, we compared \muplan, against
the \textsc{e2e} system, and  reference
summaries. Bilingual raters were shown a document, alongside two summaries and
were asked to provide pairwise references along the following
dimensions: \emph{Coherence} (is the summary easy to understand and
grammatically correct?), \emph{Accuracy} (is all the information in
the summary attributable to the original text?), and
\emph{Informativeness} (does the summary capture important information
from the original text?). We recruited~178 annotators (all
native speakers) and elicited preferences for 100 summaries (test set)
per language pair ($\textsc{en}\rightarrow \textsc{cs}$,
$\textsc{en}\rightarrow \textsc{de}$, $\textsc{en}\rightarrow
\textsc{fr}$).  Appendix~\ref{sec:human-eval-study} showcases our instructions and examples of  summaries our annotators rated. 

\begin{table}[t]
  \centering
  \small
  \begin{tabular}{@{}l@{~}rrr|rrr@{}}\toprule
&            \multicolumn{3}{c}{\muplan\ vs. \textsc{e2e}}  &
            \multicolumn{3}{c}{\muplan\ vs. Reference} \\ 
           & Win & Lose & Tie    & Win & Lose & Tie \\ \midrule
Coherence  & 6.3 & \bf 7.0 & 86.7 & \bf 10.7 & 7.6 &  81.7\\
Accuracy & \uline{\bf 13.3} & \uline{7.0} & 79.7  & \bf 15.7 & 13.6 & 70.7 \\
Inform & \uline{\bf 20.0} & \uline{11.7} & 68.3 & 14.0 & \bf 16.7 & 69.3 \\
Overall & \uline{\bf 41.0} & \uline{24.7} & 34.3 & 33.0 & \bf 35.7 & 31.3 \\\bottomrule

\end{tabular}
\caption{\label{tab:human_eval} Human evaluation results aggregated
  over three language pairs ($\textsc{en}\rightarrow \textsc{cs}$,
$\textsc{en}\rightarrow \textsc{de}$, $\textsc{en}\rightarrow
\textsc{fr}$); statistically significant differences are \uline{underlined}.}
\end{table}

We present aggregate results in Table~\ref{tab:human_eval} (see
Appendix~\ref{sec:human-eval-study} for detailed analysis).
\muplan\ summaries are as coherent as \textsc{e2e} summaries but significantly more accurate and informative ($p<0.05$ using a Wilcoxon signed-rank test). Interestingly, our raters find \muplan\ summaries on par with gold summaries across all dimensions (differences between them are \emph{not} significant).

\section{Conclusion}
In this work we present \muplan, an approach to cross-lingual
summarization that uses an intermediate planning step as a
cross-lingual bridge. Since hallucinations and mistranslations in
cross-lingual summarization are often tied to incorrect entities, we
formulate the content plan as a sequence of entities expressing
salient content and how it should be presented. Evaluation on the
XWikis dataset  demonstrates that this planning objective achieves
state-of-the-art performance in  $\textsc{en}\rightarrow \textsc{all}$
and $\textsc{all }\rightarrow \textsc{en}$ settings and  enables  zero-shot
cross-lingual transfer to new language pairs.

In
this work, we use the embedded hyperlinks in Wikipedia articles to
extract salient entities and align them on the Wikidata knowledge
base. With recent entity annotation systems such as REFINED
\cite{ayoola-etal-2022-refined}, the same operation can be applied on
out-of-domain data, including the multilingual alignment of the entity
names.  Unlike latent variable-based intermediate representations, our
content plans are interpretable (they are expressed in natural
language) and can be easily edited, e.g.,~by filtering the entities at
inference time or with a human in the loop
\cite{narayan-etal-2021-planning,narayan2022conditional,huot-etal-2023-text}. Using
forced prompting methods as described in the oracle experiments, would
also allow us to localize entity names at inference time from a
knowledge base. In the future, we  plan to explore the task transfer capabilities
of \muplan\ in low-resource settings as we cannot realistically expect to have large-scale cross-lingual data on all possible language pairs. 

\section*{Limitations}

An ethical consideration with generative language models is the problem of misinformation. While the work we present here makes a step towards improving the faithfulness and factual consistency of text generation systems, it is important to note that current systems are still far from  perfect in this respect. They can make mistakes and thus their output should be checked and used with caution.

\bibliography{anthology,custom}
\clearpage
\appendix

\section{Cross-lingual Summarization Datasets} 
\label{sec:crosslingual-datasets}

Table~\ref{tab:datasets-comparison} summarizes existing cross-lingual datasets. We see that the XWikis dataset \cite{perez-beltrachini-lapata-2021-models} features longer input documents and target summaries.

\begin{table}[t]
    \centering
    \resizebox{\columnwidth}{!}{%
    \begin{tabular}{lcccc}
    \toprule
      &     Lang &  Pairs  & SumL &  DocL \\
    \midrule
    MultiLing’13 &  40 &      30 &  185 & 4,111 \\ 
    MultiLing’15 &  38 &      30 &  233 & 4,946 \\ 
    Global Voices & 15 &     229 &   51 &   359 \\ 
    WikiLingua &    18 &  45,783 &   39 &   391 \\ 
    XWikis & 4 & 213,911 &   77 &   945 \\
    CrossSum &   45  & 22,727&  23 & 431\\
    \citet{fatima-strube-2021-novel} & 2 &  50,123 & 100  & 1,572 \\
    \bottomrule
    \end{tabular}%
    }
    \caption{Number of languages (Lang), average number of document-summary pairs (Pairs), average summary (SumL) and document (DocL) length in terms of number of tokens for different cross-lingual datasets.}
    \label{tab:datasets-comparison}
\end{table}

\section{Cross-lingual NLI}
\label{sec:nli-analysis}
Table~\ref{tab:nli} compares different ways of computing NLI. It is
computed on the summaries generated by the baseline \textsc{e2e} model on the $\textsc{en}\rightarrow \textsc{all}$ and $\textsc{all}~\rightarrow~\textsc{en}$ tasks. The first setting, denoted as ANLI, is the English setting, for which we translate the non-English document ($\textsc{all}~\rightarrow~\textsc{en}$) or summary ($\textsc{en}\rightarrow \textsc{all}$) to English and apply an NLI model trained on an English corpus. The second one is the multilingual NLI setting, which we denote as XNLI-m. For the cross-lingual language pairs, we translate the English document or summary such that both document and summary are in the same language (which is either the source or target language, depending on whether it is the $\textsc{en}\rightarrow \textsc{all}$ or $\textsc{all}~\rightarrow~\textsc{en}$ task). We then apply a multilingual NLI model. The last setting is the cross-lingual setting, which we denote as XNLI-x. In this setting, we do not use translation, and directly apply the multilingual NLI model to the cross-lingual data.

\begin{table}[t]
    \centering
    \begin{tabular}{clccc}
        \toprule
        & & ANLI  & XNLI-m    & XNLI-x  \\
        \midrule
        \multirow{4}{*}{\begin{turn}{90}$\textsc{en}\rightarrow \textsc{all}$\end{turn}}
        & \textsc{en}     & 54.04 & -- & 53.63  \\
        & $\textsc{en}\rightarrow \textsc{cs}$ & 32.09 & 31.15 & 35.88  \\
        & $\textsc{en}\rightarrow \textsc{de}$ & 38.47 & 39.89 & 40.15  \\
        & $\textsc{en}\rightarrow \textsc{fr}$ & 43.09 & 35.74 & 41.32  \\
        \midrule
        \multirow{4}{*}{\begin{turn}{90}$\textsc{all}\rightarrow \textsc{en}$\end{turn}}
        & \textsc{en}       & 57.91  & --    & 53.05  \\
        & $\textsc{cs}\rightarrow \textsc{en}$     & 34.73   & 32.95     & 29.74 \\
        & $\textsc{de}\rightarrow \textsc{en}$     & 40.28   & 38.64    & 35.12 \\
        & $\textsc{fr}\rightarrow \textsc{en}$    & 37.28   & 35.71   & 32.40 \\
        \bottomrule
    \end{tabular}
    \caption{Entailment metrics on English, multilingual, and cross-lingual settings.}
    \label{tab:nli}
\end{table} 

\section{Experimental Results}
\label{sec:more-results}

In Table~\ref{tab:results_overview} we present the full set of ROUGE
scores for the $\textsc{en}\rightarrow \textsc{all}$ and
$\textsc{all}\rightarrow \textsc{en}$ tasks.

\begin{table*}[t]
  \centering
  \small
  \begin{tabular}{l|cccc|cccc}\toprule
 \multicolumn{1}{c}{}   &   \multicolumn{4}{c}{ROUGE-1} & \multicolumn{4}{c}{ROUGE-2} \\
  & \multicolumn{1}{c}{\textsc{TR}$_{train}$} &
  \multicolumn{1}{c}{\textsc{TR}$_{test}$} &
  \multicolumn{1}{c}{\textsc{e2e}} & \multicolumn{1}{c}{\muplan}
&  \multicolumn{1}{|c}{\textsc{TR}$_{train}$} &
  \multicolumn{1}{c}{\textsc{TR}$_{test}$} &
  \multicolumn{1}{c}{\textsc{e2e}} & \multicolumn{1}{c}{\muplan}  \\ \midrule
  \small{$\textsc{en}\rightarrow \textsc{en}$}    & 45.38 & \bf 47.95 & 45.47 & 47.43 & 28.61 & 30.26 & 28.73 & \bf 30.61 \\
  \small{$\textsc{en}\rightarrow \textsc{cs}$}    & 40.74 & 35.12 & 40.72 & \bf 41.02 & 23.86 & 17.08 & 23.70 & \bf 24.43 \\
  \small{$\textsc{en}\rightarrow \textsc{de}$}    & 44.51 & 37.49 & 44.58 & \bf 45.34 & 28.99 & 18.27 & 29.26 & \bf 29.35 \\
  \small{$\textsc{en}\rightarrow \textsc{fr}$}    & 48.69 & 42.15 & 48.73 & \bf 49.23 & 32.81 & 22.00 & 32.89 & \bf 33.20 \\ \midrule
  \textbf{$\textsc{en}\rightarrow \textsc{all}$}  & 44.83 & 40.68 & 44.87 & \bf 45.75 & 28.56 & 21.90 & 28.65 & \bf 29.40 \\ \bottomrule
\multicolumn{1}{c}{}\\ \toprule
 \multicolumn{1}{c}{}   &   \multicolumn{4}{c}{ROUGE-1} & \multicolumn{4}{c}{ROUGE-2} \\
& \textsc{TR}$_{train}$ & \textsc{TR}$_{test}$ & \textsc{e2e} & \muplan
& \textsc{TR}$_{train}$ &\textsc{TR}$_{test}$ & \textsc{e2e} & \muplan
  \\ \midrule
\small{$\textsc{en}\rightarrow \textsc{en}$}   & 40.61 & 42.87 & 44.57 & \bf 44.65
& 21.12 & 25.24 & 25.61 & \bf 26.52 \\
\small{$\textsc{cs}\rightarrow \textsc{en}$}   & 36.80 & 41.46 & \bf 43.48 & 43.18
& 16.85 & 20.53 & \bf 22.46 & 22.06\\
\small{$\textsc{de}\rightarrow \textsc{en}$}   & 37.47 & 40.18 & 43.15 & \bf 43.22
& 17.32 & 21.93 & 23.38 & \bf 24.21 \\
\small{$\textsc{fr}\rightarrow \textsc{en}$}   & 36.82 & 40.83 & 42.85 & \bf 43.19
& 17.17 & 21.85 & 22.75 & \bf 23.98 \\ \midrule
\textbf{$\textsc{all}\rightarrow \textsc{en}$} & 37.93 & 41.34 & 43.51 & \bf 43.56
& 18.11 & 22.39 & 23.55 & \bf 24.19 \\ \bottomrule
\end{tabular}
    \caption{ROUGE-1 and ROUGE-2 results per language pair and overall
      for the $\textsc{en}\rightarrow \textsc{all}$ and
      $\textsc{all}\rightarrow \textsc{en}$ tasks.}
    \label{tab:results_overview}
\end{table*}

\begin{table*}[h!]
    \centering
    \begin{tabular}{l|cc|cc} \toprule
        \multicolumn{1}{c}{}    &\multicolumn{2}{c}{\textbf{$\textsc{en}\rightarrow \textsc{all}$}} &\multicolumn{2}{c}{\textbf{$\textsc{all}\rightarrow \textsc{en}$}} \\
         & ROUGE-L  & XNLI   & ROUGE-1 / 2 / L   & XNLI  \\
        \midrule

		\textsc{e2e}   & 44.54	/ 28.57	/ 37.40  & 42.75  & 43.54	/ 23.44	/ 33.79  & 37.58  \\
		\hspace{0.3cm} \footnotesize{\emph{filtered}}    & 44.87 / 28.65 / 37.56  & 41.77  & 43.51 / 23.55 / 33.92  &  37.87 \\
        \bottomrule
    \end{tabular}
    \caption{Comparison of cross-lingual summarization results obtained with \emph{filtered} and non-filtered training data.}
    \label{tab:filtered}
\end{table*}

\section{Effects of Filtered Training Data}
\label{sec:filtered}

Table \ref{tab:filtered} compares the results obtained with the \emph{filtered} and non-filtered training data. Overall, the results are similar, which is expected since the difference in the number of training samples is relatively small.

\section{Few-shot Prompting of LLMs}
\label{sec:few-shot-LLM}

\begin{table}[t]
    \centering
    \small
    \begin{tabular}{l|cc}
        \toprule
    &   ROUGE-L & XNLI \\ \midrule
  \small{$\textsc{en}\rightarrow \textsc{en}$}  &   36.37 & 36.87 \\
  \small{$\textsc{en}\rightarrow \textsc{cs}$}  &   28.64 & 31.90  \\
  \small{$\textsc{en}\rightarrow \textsc{de}$}  &   32.83 & 31.68  \\
  \small{$\textsc{en}\rightarrow \textsc{fr}$}  &  39.93 & 34.40 \\ \midrule
  \textbf{$\textsc{en}\rightarrow \textsc{all}$}  & 34.44  & 33.71 \\ \bottomrule
\multicolumn{1}{c}{}\\ \toprule
    &   ROUGE-L & XNLI \\ \midrule
\small{$\textsc{en}\rightarrow \textsc{en}$}  &   36.37 & 36.87 \\
\small{$\textsc{cs}\rightarrow \textsc{en}$}  &   26.27 & 29.00\\
\small{$\textsc{de}\rightarrow \textsc{en}$}  &   34.97 & 32.68\\
\small{$\textsc{fr}\rightarrow \textsc{en}$}  &   30.39 & 24.44 \\ \midrule
\textbf{$\textsc{all}\rightarrow \textsc{en}$}  & 32.00  & 30.75\\ \bottomrule
    \end{tabular}
    \caption{One-shot prompting results with PaLM 2 per language pair and overall
      for the $\textsc{en}\rightarrow \textsc{all}$ and
      $\textsc{all}\rightarrow \textsc{en}$ tasks.}
    \label{tab:llm}
\end{table} 

LLMs have demonstrated promising results in few-shot settings for cross-lingual summarization \cite{wang2023cross}. In Table \ref{tab:llm}, we report 1-shot results obtained using  PaLM 2 \cite{anil2023palm}, a 340B parameter LLM. 
We perform 1-shot experiments for all language pairs in the $\textsc{en}\rightarrow \textsc{all}$ and $\textsc{all}~\rightarrow~\textsc{en}$ tasks. For each language pair, the prompt is formulated as follows: 

\footnotesize{
\begin{verbatim}

From a document in [source language],
write a summary in [target language].

(1)
Document: [example document]
Summary: [example summary]

(2)
Document: [document]
Summary:

\end{verbatim}
}

\normalsize

The example document and summary are taken from the training splits. We truncate the input documents at 2000 tokens to fit within the model’s maximum sequence input length. We limit the experiments to the 1-shot setting, since more than one data example exceeds the maximum sequence length.

These 1-shot LLM experiments underperformed overall compared to our finetuned baselines. The ROUGE-L scores are lower than both the \textsc{e2e} and \muplan~models and the NLI scores are much lower than all models. In the $\textsc{en}\rightarrow \textsc{cs}$ task, the model often generated outputs in English instead of Czech. These results highlight some of the challenges of learning cross-lingual summarization from just a few examples.

While the few-shot setting has its limitations, fine-tuning large language models (LLMs) is computationally expensive, and not suited for studies with many experiments. 
% Moreoever, the pretrained language model benchmark by \citealt{maynez2023benchmarking} reported that pretrained encoder-decoder models (such as mT5 \cite{xue-etal-2021-mt5}) are better suited to summarization tasks than some of the larger pretrained language models that use a decoder-only architecture. Therefore, we use the mT5 model for our study.

\section{Human Evaluation Study}
\label{sec:human-eval-study}

\begin{figure*}[t]
    \centering
        % trim left bottom right top
        \includegraphics[width=\textwidth,trim={0 180 0 10},clip]{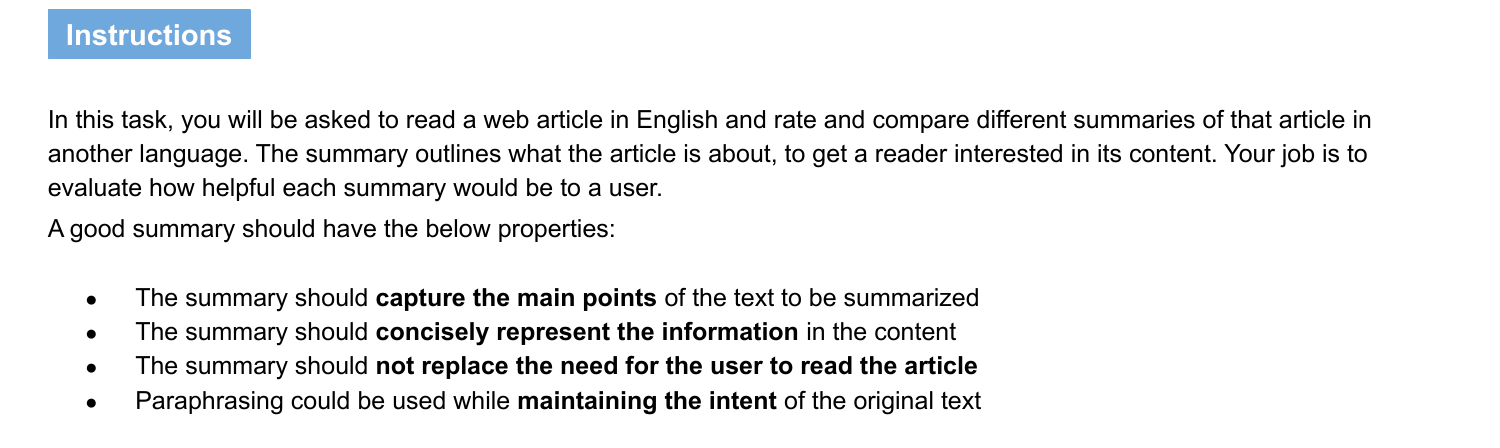}
        \includegraphics[width=\textwidth,trim={0 50 0 10},clip]{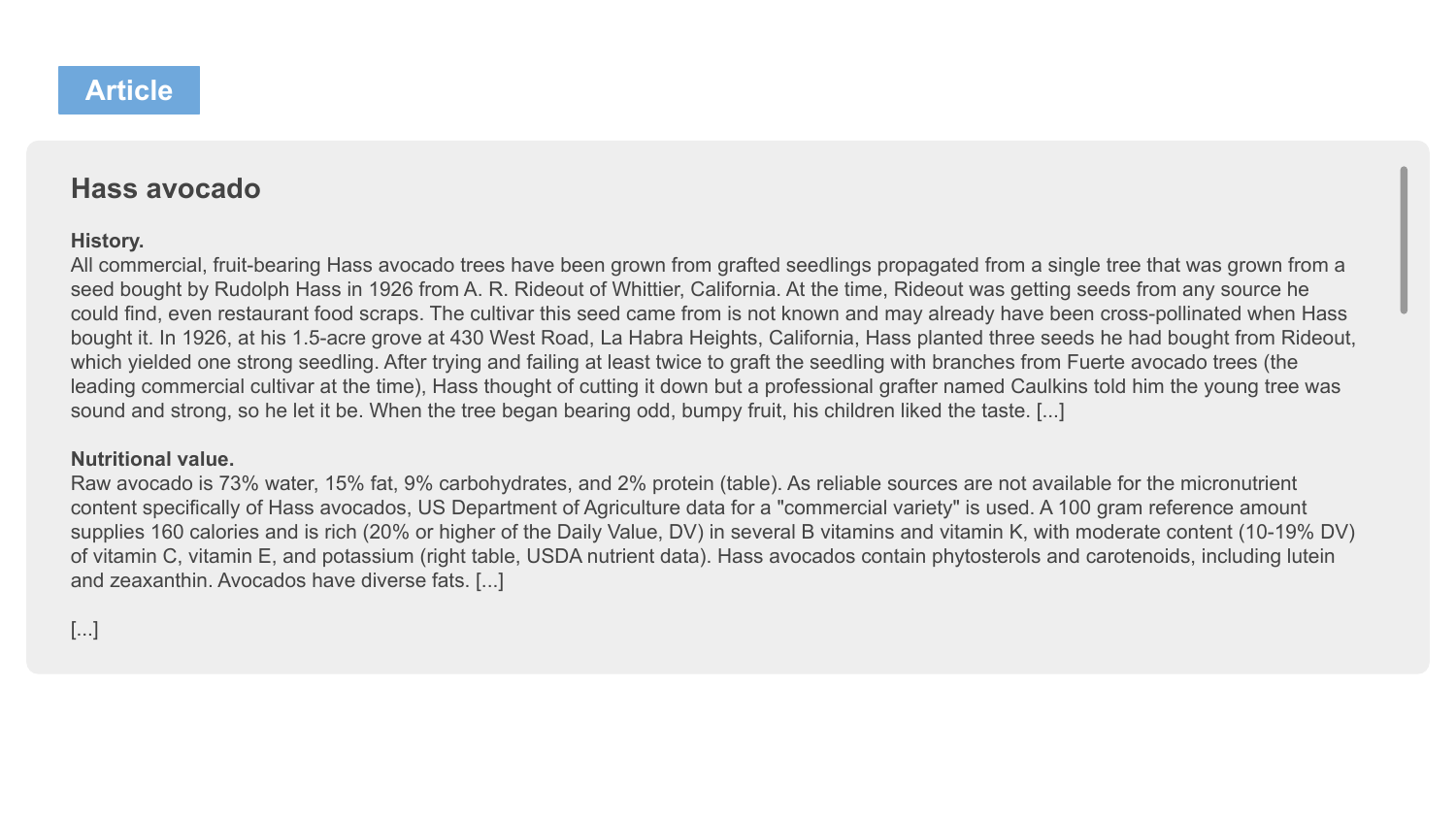}
        \includegraphics[width=\textwidth,trim={0 80 0 10},clip]{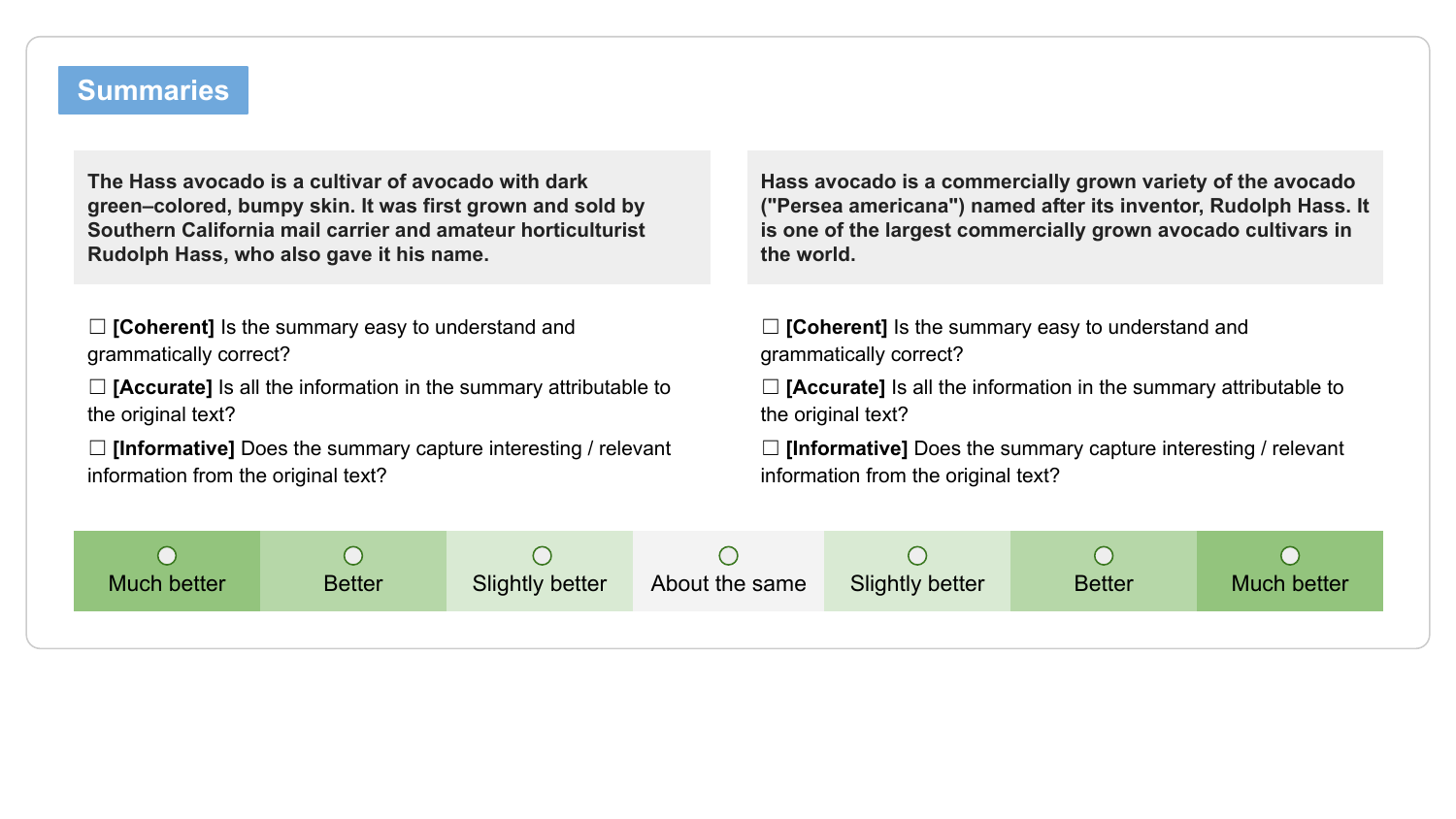}
    \caption{Experimental instructions presented to participants during our human elicitation study.}
    \label{fig:instructions}
\end{figure*}

Figure~\ref{fig:instructions} presents the experimental  instructions used in our human elicitation study.
To recruit our participants, we screened their language skills to determine whether they are native speakers, their education level and country of residence as well as origin. In addition, we created a screener test to determine the raters' suitability for the task. In total, we recruited 178 annotators across  four languages.  Our annotators were paid adequately by our suppliers adhering to the supplier code of conduct. 

Tables~\mbox{\ref{tab:German:summaries} and~\ref{tab:Czech:summaries}} show examples of the summaries rated by our participants (gold-standard references or output generated by  \muplan\ and the \textsc{e2e} systems).

\clearpage
\begin{table*}[t]
\begin{tabular}{lp{13cm}} 
\multicolumn{2}{c}{Hill of Tara (\url{https://en.wikipedia.org/wiki/Hill_of_Tara})} \\ \toprule
%\multicolumn{2}{c}{} \\ \hline
\textbf{\textsc{e2e}} & La colline de Tara (en irlandais : "Cnoc na Teamhrach", « colline des rois ») est une colline du comté de Dublin en République d'Irlande.\\ \hline
\textbf{\muplan} & La colline de Tara (en irlandais "Cnoc na Teamhrach", « colline des rois ») est une colline située à l'est de Dublin, en Irlande. C'était autrefois le haut lieu des rois d'Irlande. Dans la mythologie celtique irlandaise, elle était la capitale des Tuatha Dé Danann.\\ \hline
\textbf{Reference} & Tara est un site archéologique d'Irlande dans le comté de Meath. Dans la mythologie celtique irlandaise, Tara est la capitale mythique de l'Irlande, située dans la cinquième province de Mide, dans le centre du pays : c'est la « colline des rois » ().\\\bottomrule
\multicolumn{2}{c}{} \\ 
\multicolumn{2}{c}{Richelsley (\url{https://en.wikipedia.org/wiki/Richelsley})} \\ \toprule
\textbf{\textsc{e2e}} & Le Richelsley est un rocher de l'Eifel, situé dans la commune de Waimes, près de Reichenstein/Monschau, en Belgique.
\\ \hline
\textbf{\muplan} & 
Le Richelsley est une formation rocheuse située dans la commune de Waimes, dans la province de Liège, en Région wallonne, en Belgique, près de la frontière allemande, à proximité de l'abbaye de Reichenstein/Monschau. Le Richelsley est surtout connu pour sa grande croix de six mètres de haut, qui a été érigée en 1890 par le prêtre Gerhard Joseph Arnoldy, qui travaillait de 1869 à 1914 à Kalterherberg et était le bâtisseur de l'actuelle cathédrale d'Eifel. Le roman "Das Kreuz im Venn" de Clara Viebig a fait du Richelsley un lieu de pèlerinage.\\ \hline
\textbf{Reference} & Le rocher de Richelsley est un important rocher isolé situé à l'est de la Belgique dans les Hautes Fagnes et faisant partie de la commune de Waimes. Sur ce rocher, se dresse une croix appelée Kreuz im Venn. \\ 
\bottomrule
\end{tabular}
\caption{\label{tab:French:summaries} Examples of system output (\textsc{e2e}, \muplan) and gold-standard reference for the  $\textsc{en}\rightarrow \textsc{fr}$ setting. Only title and url are shown for input Wikipedia article, for the sake of brevity.}
\end{table*}

\begin{table*}[t]
\begin{tabular}{lp{13cm}} 
\multicolumn{2}{c}{Carduus (\url{https://en.wikipedia.org/wiki/Carduus}} \\ \toprule
\textbf{\textsc{e2e}} &  
Die Carduonen ("Carduus") sind eine Pflanzengattung in der Familie der Korbblütler (Asteraceae). Die etwa 90 bis 127 Arten sind fast weltweit verbreitet.\\\hline
\textbf{\muplan} &  Die Stiele ("Carduus") sind eine Pflanzengattung in der Unterfamilie Carduoideae innerhalb der Familie der Korbblütler (Asteraceae). Die etwa 90 bis 127 Arten sind in den gemäßigten Gebieten der Nordhal.\\\hline
\textbf{Reference} & 
Die Ringdisteln ("Carduus") sind eine Pflanzengattung in der Familie der Korbblütler (Asteraceae). Die etwa 90 Arten sind ursprünglich in Eurasien und Afrika verbreitet.\\ \bottomrule 
\multicolumn{2}{c}{} \\ 
\multicolumn{2}{c}{Francesco Satolli (\url{https://en.wikipedia.org/wiki/Francesco_Satolli})} \\ \toprule
\textbf{\textsc{e2e}} & Francesco Kardinal Satolli (* 21. Juli 1839 in Marsciano, Provinz Perugia, Italien; † 8. Januar 1910 in Rom) war ein Kardinal der Römischen Kirche. \\
\hline
\textbf{\muplan} & Francesco Kardinal Satolli (* 21. Juli 1839 in Marsciano; † 8. Januar 1910 in Rom) war ein italienischer Geistlicher und Kurienkardinal.\\
\hline
\textbf{Reference} & 
Francesco di Paola Kardinal Satolli (* 21. Juli 1839 im Marsciano in der Nähe von Perugia; † 8. Januar 1910 in Rom) war ein italienischer Theologe, Kardinal und der erste Apostolische Delegat in die Vereinigten Staaten.
\\  \bottomrule
\end{tabular}
\caption{\label{tab:German:summaries} Examples of system output (\textsc{e2e}, \muplan) and gold-standard reference for the $\textsc{en}\rightarrow \textsc{de}$ setting. Only title and url are shown for input Wikipedia article, for the sake of brevity.}
\end{table*}

\begin{table*}[t]
\begin{tabular}{lp{13cm}} 
\multicolumn{2}{c}{Legendre's formula  (\url{https://en.wikipedia.org/wiki/Legendre\%27s\_formula})} \\ \toprule
%\multicolumn{2}{c}{} \\ \hline
\textbf{\textsc{e2e}} & 
Legendreova věta je matematické tvrzení z oblasti teorie čísel, které umožňuje počítat součty čísel a rozdělení jejich mocnin: Tedy například pokud "n" je kladné a "p" je velké číslo, pak
\\\hline
%\multicolumn{2}{c}{} \\ \hline
\textbf{\muplan} & Legendrova věta je matematické tvrzení z oblasti teorie čísel, které říká, že pro každé prvočíslo "p" a každé přirozené číslo "n" existuje největší mocninná mocnina "p", která dělí "n". Tedy
\\ \hline
\textbf{Reference} & Legendreův vzorec (také De Polignacův vzorec) dovoluje vypočítat nejvyšší exponent u prvočísla formula\_1, kde formula$\_$1 umocněné na tento exponent ještě dělí číslo formula$\_$3 (faktoriál přirozeného čísla formula$\_$4). Jedná se v podstatě o výpočet p-adické valuace čísla formula$\_$3.
\\\bottomrule
\multicolumn{2}{c}{} \\ 
\multicolumn{2}{c}{Stack overflow (\url{https://en.wikipedia.org/wiki/Stack_overflow})} \\ \toprule
\textbf{\textsc{e2e}} & Přetečení na zásobníku () je v informatice označení pro speciální případ přetečení na zásobníku. Přetečení na zásobníku nastává v důsledku umělého vyčerpání úložného prostoru programem. V důsledku toho dojde k přepsání obsahu na zásobník, což způsobí vznik segmentation fault.
 \\\hline
\textbf{\muplan} &  Přetečení na zásobníku () je v informatice situace, kdy je v počítači program, který používá zásobník operační paměti, více než je v ní uložen. Přetečení na zásobníku vzniká většinou v důsledku dlouhých procesů, které konají nadbytečné údaje. Nejčastější příčinou přetečení na zásobníku je nekonečná rekurze. V programech s více vlákny je přetečení na zásobníku běžné.
\\\hline
\textbf{Reference} & Přetečení zásobníku je situace v programování, kdy dojde k pokusu uložit na zásobník volání více dat, než kolik se tam vejde. Velikost tohoto zásobníku je obvykle předem dána při startu programu v závislosti na architektuře systému, překladači, množství volné paměti atp. Když se program pokusí posunout vrchol zásobníku mimo vymezenou paměť, mluvíme o přetečení zásobníku. To má obvykle za následek pád programu.\\
\bottomrule
\end{tabular}
\caption{\label{tab:Czech:summaries} Examples of system output (\textsc{e2e}, \muplan) and gold-standard reference for the  $\textsc{en}\rightarrow \textsc{cz}$ setting. Only title and url are shown for input Wikipedia article, for the sake of brevity.}
\end{table*}

\end{document}